\documentclass[letterpaper, 10 pt, conference]{ieeeconf}  

\IEEEoverridecommandlockouts                              
\usepackage{graphicx}
\usepackage{url}
\usepackage{listings}
\usepackage{courier} 
\usepackage{color}
\usepackage{textcomp}

\definecolor{gray}{rgb}{0.8,0.8,0.8}

\lstdefinelanguage{Lua}{
  morekeywords={
    and,break,do,else,elseif,end,false,for,function,if,in,
    local,nil,not,or,repeat,return,then,true,until,while,
  },
  sensitive=true,
  morecomment=[l]{--},
  backgroundcolor=\color{gray},
  showspaces=false,               
  showstringspaces=false,         
  showtabs=false,                 
  basicstyle=\footnotesize \ttfamily,
  captionpos=b,
  aboveskip=\bigskipamount, 
  basicstyle=\scriptsize \ttfamily,
  upquote=true,
  aboveskip=\bigskipamount, 
  columns=fixed,
  showstringspaces=false,
  extendedchars=true,
  breaklines=true,
  prebreak =,
  rulecolor=,
  inputencoding=utf8x,
  morecomment=[n]{--[[}{]]},
  morestring=[b]'',
  morestring=[b]',
}[keywords,comments,strings]

\title{\LARGE \bf
  Pure Coordination using the Coordinator--Configurator Pattern
}

\author{Markus Klotzb\"ucher, Geoffrey Biggs and Herman Bruyninckx
\thanks{Markus Klotzb\"ucher and Herman Bruyninckx are with the Department of Mechanical Engineering,
  KU Leuven, Belgium. Geoffrey Biggs is at the Intelligent Systems Research
  Institute, AIST, Japan. Corresponding author: {\tt\small
      markus.klotzbuecher@mech.kuleuven.be}}
\thanks{This research was funded by the European Community under grant
  agreements FP7-ICT-231940 (Best Practice in Robotics), and
FP7-ICT-230902(\emph{ROSETTA}), and by KU Leuven's Concerted Research
Action \emph{Global real-time optimal control of autonomous robots and
mechatronic systems}.}}

\begin{document}

\maketitle
\thispagestyle{empty}
\pagestyle{empty}

\begin{abstract}
  This work-in-progress paper reports on our efforts to
  improve different aspects of \textit{coordination} in complex,
  component-based robotic systems.  Coordination is a system level
  aspect concerned with commanding, configuring and monitoring
  functional computations such that the system as a whole behaves as
  desired. To that end a variety of models such as Petri-nets or
  Finite State Machines may be utilized. These models
  specify actions to be executed, such as \textit{invoking} operations
  or configuring components to achieve a certain goal.

  This traditional approach has several disadvantages related to loss
  of reusability of coordination models due to coupling with platform-specific
  functionality, non-deterministic temporal behavior and
  limited robustness as a result of executing platform operations
  within the context of the coordinator.

  To avoid these shortcomings, we propose to split this
  ``rich'' coordinator into a \textit{Pure Coordinator} and a
  \textit{Configurator}. Although the coordinator remains in charge of
  commanding and reacting, the execution of actions is deferred to the
  Configurator. This pattern, called
  \textit{Coordinator--Configurator}, is implemented as a novel
  Configurator domain specific language that can be used together with
  any model of coordination. We illustrate the approach by refactoring
  an existing application that realizes a safe haptic coupling of two
  youBot mobile manipulators.
\end{abstract}

\section{Introduction}

The context of this work (and hence also for the described pattern)
is complex, component-based robotics and machine tool systems
operating under real-time constraints. For building such systems, an
increasingly acknowledged best practice is to separate the concerns of
Coordination, Computation, Configuration and Communication
\cite{RadestockEisenbach1996,prasslerbruyninckxnilssonshakhimardanov2009}.

Computation defines the basic, functional building blocks from which a
system is constructed. Communication defines how and with whom the
individual elements of a system communicate. Configuration defines the
properties of a system . Lastly, Coordination is concerned with
supervising and monitoring the computations in way that the system as
a whole behaves as intended.

Classical coordination models that have been used in robotics are
Petri-Nets \cite{Rosell2004}, Finite State Machines (FSM)
\cite{FinkemeyerKroegerWahl2005} and Statecharts
\cite{marty1998,Klotzbuecher2011-IROS}. These models are used to
define when certain behaviors shall be executed. For instance UML FSM
\cite{uml-superspec2011} allows execution of a behavior upon entering or
exiting a state. The exact actions available depend on the primitives
of the underlying framework, and may include \textit{invoking}
operations or modifying the configuration of a component.

This traditional approach has three major disadvantages. Firstly,
reusability of coordination models is reduced because the model is
polluted with platform specific information. In other words, reusing
the same model on a different robot or software framework requires
intrusive refactoring to replace the platform specific operations used
in coordinator. Secondly, the blocking invocation of operations on
functional computations can severely degrade the determinism of the
Coordinator. Lastly, Coordinator robustness is reduced since an
invocation might block indefinitely or crash, either of which may
effectively render the coordinator inoperative.

\begin{figure}[h]
  \centering
  \includegraphics[width=0.9\columnwidth]{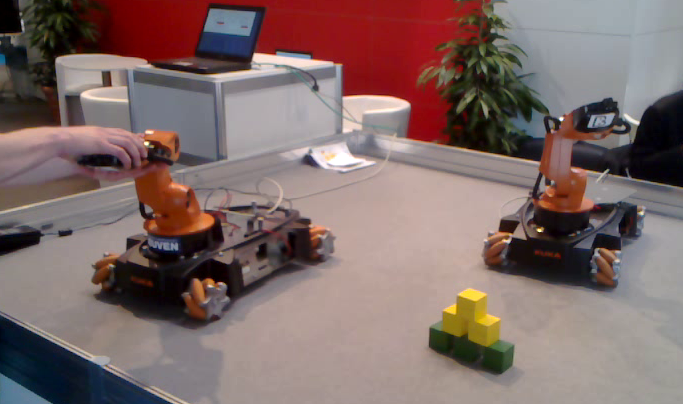}
  \caption{Bidirectional youBot coupling demonstration: each robot
    copies the cartesian position of its peer robot.}
  \label{fig:demo_screenshot}
\end{figure}

This paper proposes the \textit{Coordinator--Configurator} pattern to
overcome these challenges. The Configurator has been implemented as an
Lua \cite{lua-url} based internal domain specific language (DSL) for
the Orocos Real Time Toolkit (RTT) framework \cite{soetens2006} and is
to be complemented by a Coordination model such as rFSM
\cite{rfsm-url}. We have applied a preliminary version to a moderately
complex application\footnote{This demo was shown at the Automatica
  trade fair 2012 in Munich. } consisting of a haptic,
force-controlled coupling of two KUKA youBots in which multiple
constraints are monitored and their violation is reacted to by the
coordinator.

\subsection{Related work}
\label{sec:related-work}

From a object oriented software engineering perspective, the classical
\textit{command} design pattern \cite{gamma-etal95} comes close to our
suggestion by permitting a \textit{client} to request an
\textit{invoker} to execute a given \textit{command} on a
\textit{recipient}.

The ROS \cite{quigley2009} framework provides the actionlib library as
a standardized protocol to define commands that can be executed,
monitored, aborted, etc. Hence it is not a form of Configurator, but
rather a mechanism that could be used to implement one.

Since this work is concerned with constructing modular subsystems, the
work of the Ptolemy project \cite{Eker2003} is relevant, although that is more
focused on composition of heterogeneous systems.

\subsection{Outline}

The remainder of this article is structured as follows. The following
section describes the Coordinator--Configurator
pattern in detail and introduces the configuration DSL that underlies
the Configurator. Section~\ref{sec:discussion} critically examines the
solution and discusses further potential uses of the
DSL. Section~\ref{sec:conclusion} concludes and describes future work.

\section{Approach}
\label{sec:approach}

\begin{figure}
  \centering
  \includegraphics[width=0.65\columnwidth]{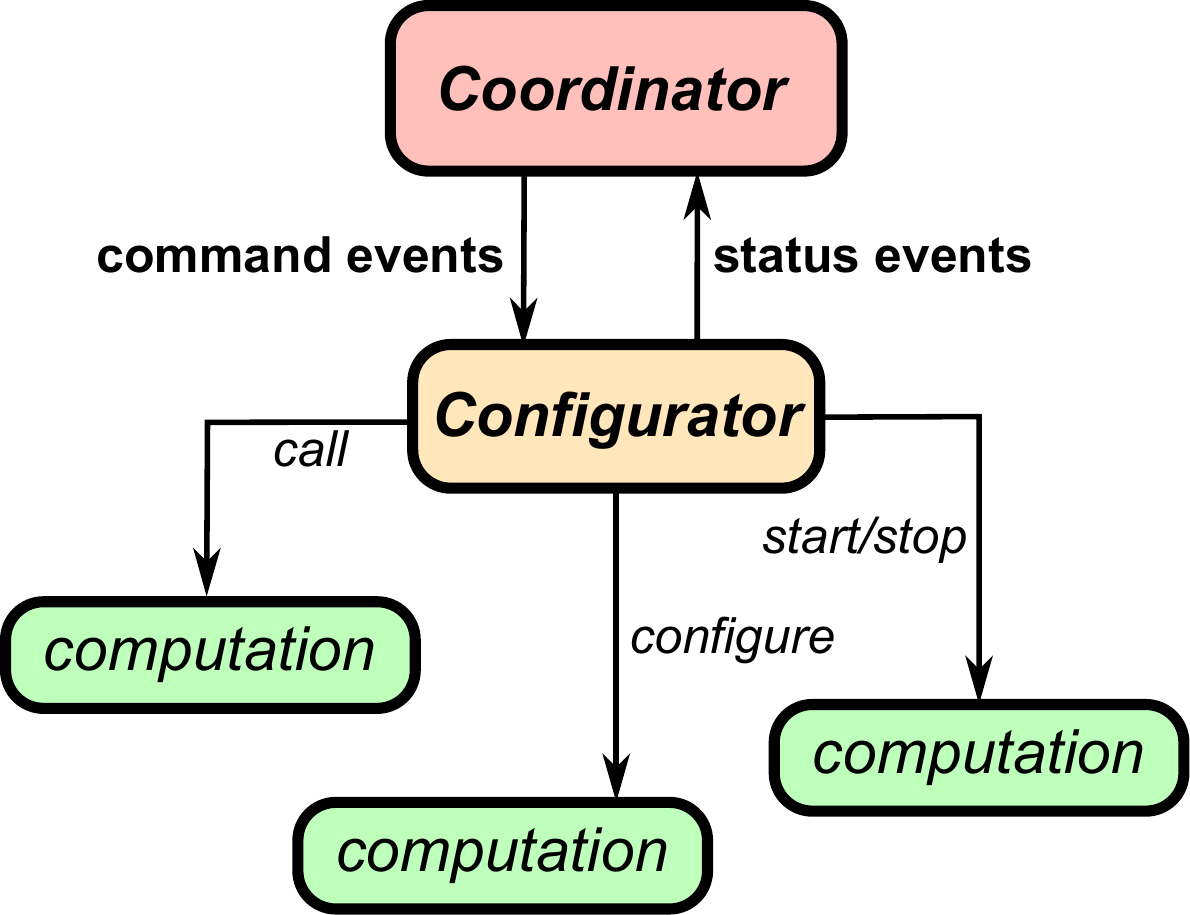}
  \caption{The relationship between the Coordinator and the Configurator.}
  \label{fig:coord_conf_relationship}
\end{figure}

To overcome the described shortcomings, we propose to split the
``rich'' coordinator into a \textit{Pure Coordinator} and a
\textit{Configurator}, named the \textit{Coordinator--Configurator}
pattern. Although the coordinator remains in charge of commanding and
monitoring, the execution of actions is deferred to the Configurator.
The Configurator is realized as a software entity (typically a
component), that is configured with a \textit{set of
  configurations}. Each configuration describes one possible state of
the system and is identified by a unique name. Furthermore, a
Configurator has a mechanism to receive events. When an event that
matches the ID of a configuration is received, the Configurator
\textit{applies} the respective configuration (details on this
application follow below). Success or failure is reported back via a
status event, permitting the Coordinator to react
appropriately. Figure~\ref{fig:coord_conf_relationship} illustrates
this mechanism.

Note that Figure~\ref{fig:coord_conf_relationship} omits the important but complementary concept of
a monitor, which is responsible for observing the system and
generating events when certain conditions are met or violated.

\section{Example}
\label{sec:Example}

\begin{figure}
  \centering
  \includegraphics[width=0.7\columnwidth]{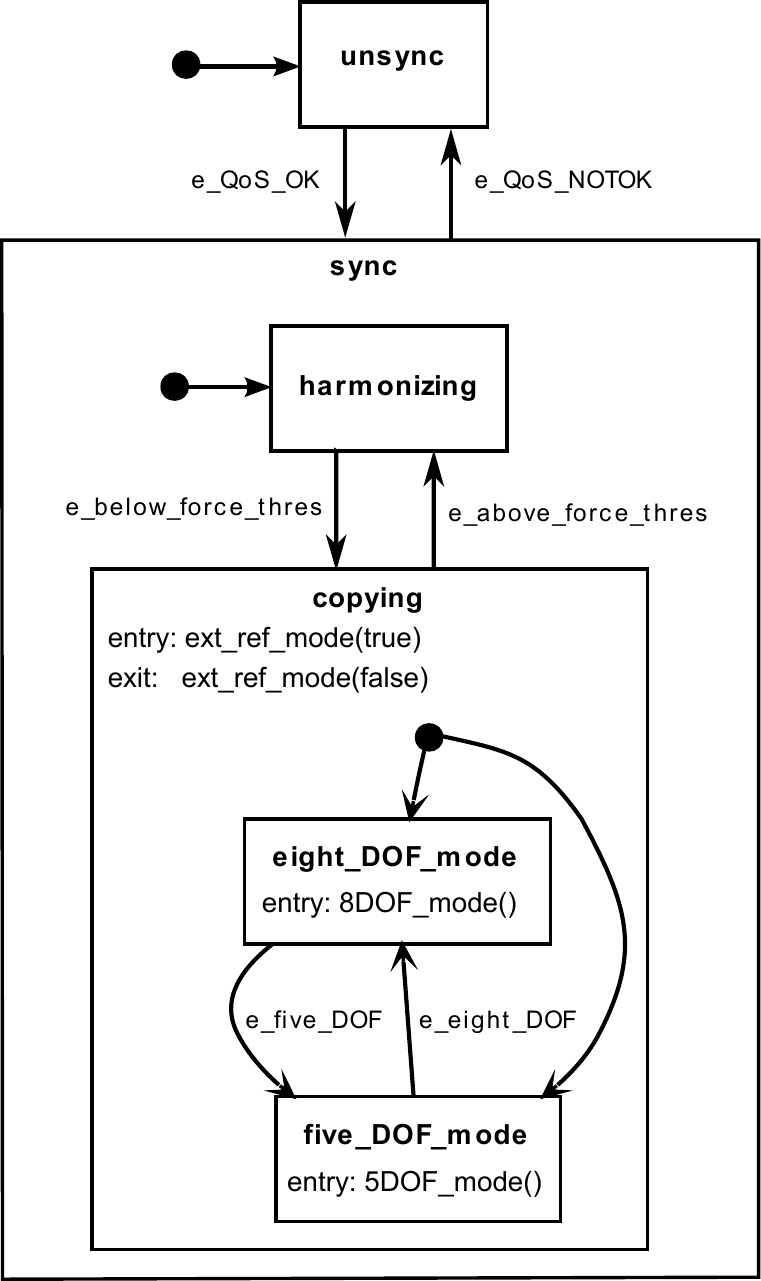}
  \caption{rFSM Coordination Statechart for the youBot coupling demo.}
  \label{fig:youbot_coordinator}
\end{figure}

\begin{figure}
  \centering
  \includegraphics[width=\columnwidth]{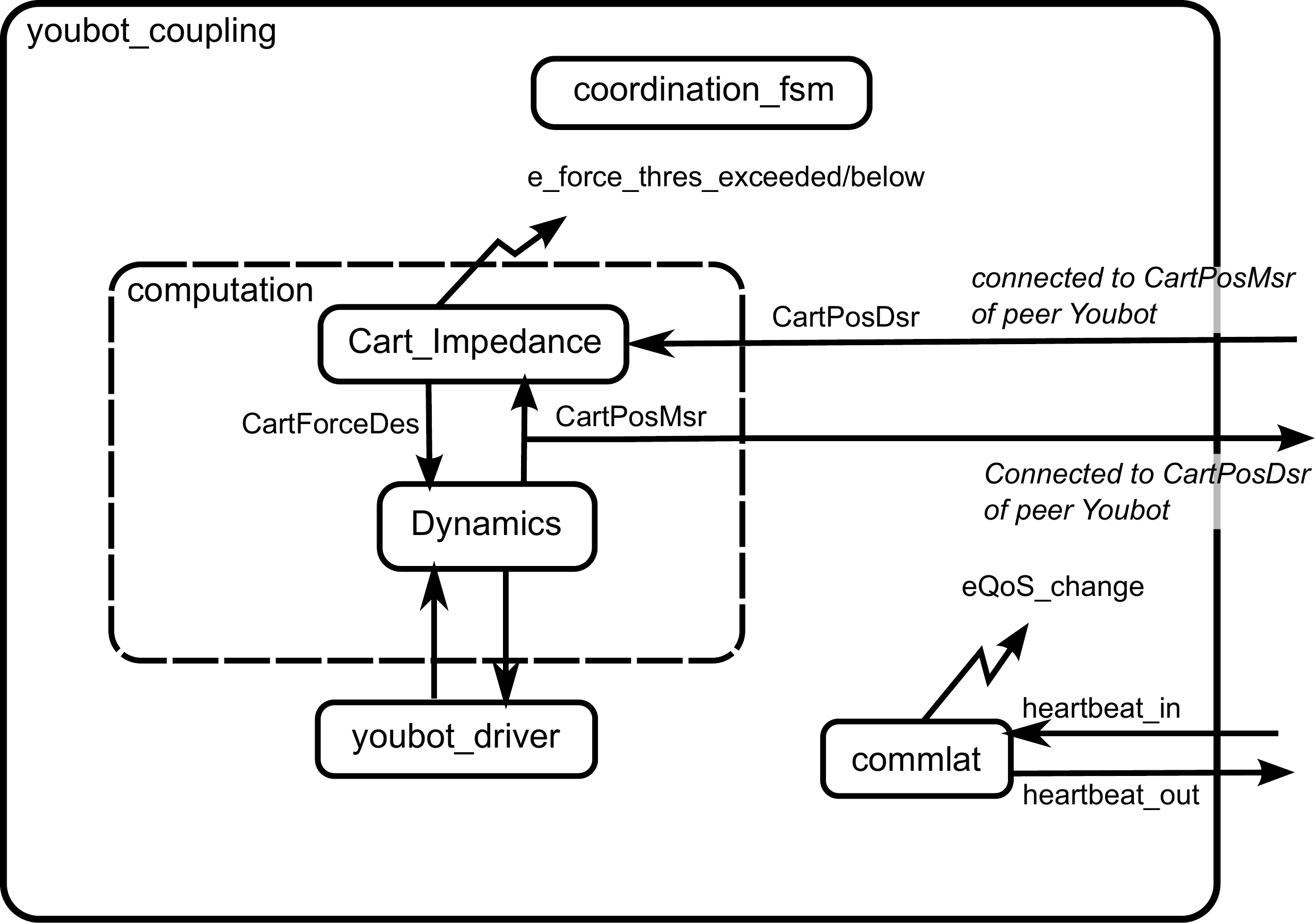}
  \caption{Component diagram illustrating the components and
    connections running on each of the youBots in the example.}
  \label{fig:component_diagram}
\end{figure}

Figure~\ref{fig:youbot_coordinator} shows the coordination statechart
that is executed on each of the two youBots of the coupling
application. Figure~\ref{fig:component_diagram} shows a (slightly
simplified) component architecture; straight arrows represent data-flow
communication and zigzag lines represent status events emitted by
computational components and received by the coordinator. This
application is used as a running example throughout this article. It has been
converted to the Coordinator--Configurator pattern.

The basic behavior of the demo is immediately visible from the
statechart. At the toplevel, the \texttt{unsync} (unsynchronized)
state is entered by default and signifies that communication with the
peer robot is not functional. After communication is established
and its QoS is sufficient, a transition to the \texttt{sync}
(synchronized) state takes place, and from there to the
\texttt{harmonizing} state. The latter means that the control loop
responsible for moving the end-effector of the robot towards the
position of the peer robot is operational and the impedance controller
\texttt{Cart\_Impedance} (see Figure~\ref{fig:component_diagram}) is
computing a desired-force control signal. However, since this force is
too high, this controller is configured to not output the signal but
instead output a desired force of zero in all directions
(\texttt{ext\_ref\_mode(false)}. As a result, the robot arm remains
controlled in a compliant way, merely compensating for gravity. Only
when both arms are (more or less) manually aligned by the operator does
the desired force fall beneath the threshold, and thus trigger the
transition to the \texttt{copying} state and from there (depending on
the configuration) to the \texttt{eight\_DOF\_mode} or
\texttt{five\_DOF\_mode}. In either case, the coupling is put into
effect by requesting the controller to output the control signal
\texttt{ext\_ref\_mode(true)}. Alternating between
\texttt{eight\_DOF\_mode} and \texttt{five\_DOF\_mode} is commanded
by the human operator, and reconfigures the \texttt{Dynamics} component
to use the holonomic base as additional degrees of freedom or not.

\section{Modeling configuration and its application}
\label{sec:conf_representation}

The main question is how to model a configuration and what the
semantics of applying it are. Obviously, a configuration has to be able to
express the necessary platform-specific changes required for runtime
coordination. For the youBot coupling example, the following Orocos
RTT-specific primitives are sufficient:

\begin{itemize}
\item Changing the state of a component (e.g. from \texttt{running} to
  \texttt{stopped})
\item Modifying a property of a component
\item Writing a value on a port.
\end{itemize}

One of the most fundamental design choices is whether to choose a
declarative or procedural model to express the behavior of applying
these constraints. In a classical coordination model like FSM, the
configuration applied when entering one Coordinator state is typically
defined in a procedural fashion by using a function that executes several
statements. However, in most cases this approach constrains the
execution more than is necessary, as there often only exists a
partial ordering requirement of the execution of these
statements. This accidental introduction of constraints is
undesirable, since it obscures the true requirements of the system and
hinders maintenance. Thus, in our approach we opted for a purely
declarative model of a configuration (apart from a minor deviation
described below). This purely declarative approach becomes possible
because, outside of the scope of the Configurator, the Coordinator can
express ordering requirements by coordinating the Configurator to
apply a series of configurations.

Listing~\ref{lst:sample_conf} shows a single sample configuration
written in the Lua based Configurator DSL. A configuration consists of
a \texttt{pre\_conf\_state} and a \texttt{post\_conf\_state}
specification and a list of configuration changes. The first
two define to which (runtime) state components shall be brought before
and after the actual configuration takes place, while the latter list
defines the exact (platform specific) changes to be applied. Note that
with respect to ordering of configuration application, the only
guarantee made is the following: the run-time states of the components
mentioned in \textit{pre\_conf\_state} and \textit{post\_conf\_state}
are set accordingly and in the defined order before resp. after the
list of changes is applied. However, no assumptions can be made about
the order of applying the individual changes themselves. The
\texttt{\_default} keyword permits changing the state of all
components that have not been mentioned otherwise. If no
\texttt{\_default} statement is provided, the state of the unmentioned
components is not changed.

\begin{lstlisting}[language=Lua,caption={A sample configuration.},label={lst:sample_conf},float]
Configuration {
    pre_conf_state = { 'compA:running', 'compB:configure',
                       '_default:stopped' },
  
    post_conf_state = { _default='running' },
    
    property_set("compA.prop1", { 2.3, 3.4, 5.34 } ),
    port_write("compB.portX", 33.4),
    operation_call("compG.op1", arg1, arg2,...),
}
\end{lstlisting}

Using this DSL, the system configurations required by the coordinator
of Figure~\ref{fig:youbot_coordinator} can be modeled as shown in
Listing~\ref{lst:youbot_coupling_conf}.

\begin{lstlisting}[language=Lua,caption={Named configurations for the youBot sample.},label={lst:youbot_coupling_conf},float]
ConfiguratorConf {
   disable_copying = Configuration{
       port_write("Cart_Impedance.ext_ref_mode", false)
   },
   
   enable_copying = Configuration {
       port_write("Cart_Impedance.ext_ref_mode", true)
   },

   eight_DOF = Configuration {
      property_set("Dynamics.force_gain", {0.1, 0.1, 0.1})
   },

   five_DOF = Configuration {
      property_set("Dynamics.force_gain", {0, 0, 0})
   },
}
\end{lstlisting}

Since the state changes from \texttt{unsync} to \texttt{harmonizing}
do not involve any actions, but merely model the constraints that must
be satisfied before the coupling can be activated, these are not
visible in the Configurator configuration. The only configurations
necessary are for enabling and disabling the coupling (hence to be
applied in \texttt{entry} and \texttt{exit} of copying respectively)
and for switching between eight and five degrees of freedom.

%


\section{Discussion}
\label{sec:discussion}

We have implemented the described DSL and the associated configurator
for the Orocos RTT framework. The existing youBot coupling
coordination has been refactored to make use of the new
Configurator. This approach has solved the
shortcomings of the traditional approach: firstly, the coordination
model remains free of any software platform-specific actions and can
be reused with any other framework, assuming a Configurator and
corresponding configuration. Secondly, the actual changes are applied
by the Configurator while the Coordinator remains reactive and free to
deal with any other situation that may arise. Lastly, failures
within the Configurator are isolated from the Coordinator,
permitting it to react to the absence of a status event and ultimately
improving its robustness. Naturally, this robustness depends on the
run-time context of both entities and the worst-case failures
possible.

Since the Configurator constitutes an additional level of indirection,
the question of the overhead introduced is justified. Obviously, this
cannot be answered in general but will mostly depend on the level of
distribution between Coordinator, Configurator and configured
components. Nevertheless, the Coordinator--Configurator pattern offers
an advantage with respect to benchmarking and profiling the
Coordination behavior: since the execution of configuration
application is localized within a single component, the respective
measurements need likewise only to be added once. This avoids
the scattering of profiling code across the Coordination model.

Lastly, having a flat map of configurations might be suboptimal in
some cases, as for instance when it is necessary to undo a
configuration to return to a previous one. With the current model it
is necessary to manually specify the \textit{inverse} configuration,
as for instance is the case for enabling and disabling the coupling in
Listing~\ref{lst:youbot_coupling_conf}. One solution to this could be
to use a stack of configurations onto which changes can be pushed
(applied) and popped (undone) again.

\subsection{Deployment}
\label{sec:deployment}

Interestingly, the Coordinator--Configurator pattern allows dealing
with deployment as a special case of Coordination and
Configuration. To that end, the only requirement is to extend the set
of configuration actions with the following primitives:

\begin{itemize}
\item Creating components
\item Destructing components
\item Creating connections between components
\item Removing connections between components.
\end{itemize}

That way, deployment can be viewed as coordinating the system through
a series of configurations that culminates with the system having
reached an initial operational state. Likewise, shutting down the
system can be defined as applying a configuration that stops all
components followed by one resulting in their destruction. A system,
including its rules for deployment, starting up, runtime changes, and
shutdown can thus be specified in terms of a single coordination
model (which can, itself, be composed from multiple coordination models) and a
platform specific Configurator configuration.


\subsection{Composition}
\label{sec:compositionality}

The approach promises to greatly facilitate composition of systems
from systems. Any valid pair of Coordinator and platform-specific
Configurator configuration can be treated as a subsystem (sometimes
called a composite component) that can be used as a building block in a
larger system. Nevertheless, for this to work, several questions need
to be answered, notably including: what is the interface a subsystem offers, how
can the contained coordination model be controlled from the
``outside'', and how the controllable transitions can be specified.
Answering these questions is outside the scope of this paper.

\section{Conclusions}
\label{sec:conclusion}

We have described the Coordinator--Configurator pattern that is
applicable to complex component-based robot systems. The pattern's
goal is to balance the forces between increased reusability, temporal
determinism and robustness on the one hand and simplicity of the
Coordinator on the other. The key idea is to separate the
responsibility for commanding actions from the responsibility for
executing them. While the first remains with the Coordinator, the latter
is assigned to a new entity called the Configurator. The idea has been
implemented as a Configurator DSL for the Orocos RTT framework.

For future work we intend to focus on adding a complementary DSL to
describe the monitor, whose description was omitted in this work, to
further explore the outlined relationship between
Coordination/Configuration and deployment, and to validate the
hypothesis that this pattern greatly facilitates specifying platform-independent
coordinators.

\addtolength{\textheight}{-12cm}  








\bibliographystyle{abbrv}
\bibliography{actsens}

\end{document}